%% file: main.tex
\documentclass[letterpaper, 10 pt, conference]{ieeeconf}  

\IEEEoverridecommandlockouts                              

\usepackage{algorithm}
\usepackage{algpseudocode}
\usepackage[dvipsnames]{xcolor}
\usepackage{annotate-equations}
\usepackage{cancel}
\usepackage{color}
\usepackage{hyperref}

\usepackage{booktabs}
\usepackage{amsfonts}
\usepackage{graphicx}
\usepackage{hyperref}
\usepackage{textcomp}
\usepackage{listings}
\usepackage{adjustbox}
\usepackage{xspace}    
\usepackage{multirow}
\usepackage{multicol}
\usepackage{tcolorbox}
\usepackage{wrapfig}
\usepackage{diagbox}
\usepackage{subcaption}
\usepackage{amsmath}
\usepackage{pifont}
\usepackage{breakcites}
\usepackage{array}

\title{\LARGE \bf
 PIMbot: Policy and Incentive Manipulation for Multi-Robot Reinforcement Learning in Social Dilemmas  
}

\author{Shahab Nikkhoo$^{1}$, Zexin Li$^{1}$, Aritra Samanta$^{1}$, Yufei Li$^{1}$ and Cong Liu$^{1}$
\thanks{$^{1}$University of California, Riverside}%
}

\begin{document}

\maketitle
\thispagestyle{empty}
\pagestyle{empty}

\begin{abstract}
\input{body/abstract.tex}
\end{abstract}

\section{INTRODUCTION}

\input{body/Intro.tex}

\section{BACKGROUND}
\input{body/background.tex}

\section{DESIGN}
\input{body/design.tex}

\section{EVALUATION}

\input{body/eval.tex}

\section{DISCUSSION $\And$ FUTURE WORKS}
\input{body/discussion}
\section{CONCLUSIONS}
 \input{body/conclusion.tex}

\section{ACKNOWLEDGMENT}
    \input{body/ack.tex}






\bibliographystyle{IEEEtran}
\bibliography{ref}

\end{document}

%% file: body/abstract.tex
Recent research has demonstrated the potential of reinforcement learning (RL) in enabling effective multi-robot collaboration, particularly in social dilemmas where robots face a trade-off between self-interests and collective benefits. However, environmental factors such as miscommunication and adversarial robots can impact cooperation, making it crucial to explore how multi-robot communication can be manipulated to achieve different outcomes. This paper presents a novel approach, namely PIMbot, to manipulating the reward function in multi-robot collaboration through two distinct forms of manipulation: policy and incentive manipulation. Our work introduces a new angle for manipulation in recent multi-agent RL social dilemmas that utilize a unique reward function for incentivization. By utilizing our proposed PIMbot mechanisms, a robot is able to manipulate the social dilemma environment effectively. PIMbot has the potential for both positive and negative impacts on the task outcome, where positive impacts lead to faster convergence to the global optimum and maximized rewards for any chosen robot. Conversely, negative impacts can have a detrimental effect on the overall task performance. We present comprehensive experimental results that demonstrate the effectiveness of our proposed methods in the Gazebo-simulated multi-robot environment. Our work provides insights into how inter-robot communication can be manipulated and has implications for various robotic applications. 

%% file: body/Intro.tex
Robots have become ubiquitous in various applications ranging from pick-and-place operations in manufacturing to facilitating the detection of disorders in healthcare settings.~\cite{ahangar2019design, su15107790}. However, challenges remain in working in complex and unpredictable environments. Multi-robot collaboration offers a promising approach to tackling these challenges. By allowing robots to work together and share information, we can create more flexible, adaptable, and effective systems.

Recent research has demonstrated that reinforcement learning (RL) which is a type of machine learning technique, is particularly well-suited for multi-robot  collaboration~\cite{TereshchukSBPDB19,AgrawalABM22iros,GaoWZYWXWLXG22iros,ZhangQQXWZWC0ZL22iros}. RL enables robots to learn from their experiences and optimize their performance even in highly complex and dynamic environments. Utilizing RL algorithms to coordinate the actions of multiple robots can enable sophisticated systems to tackle more intricate problems than those that can be handled by a single robot alone.
One particular area where multi-robot collaboration through RL can have a significant impact is space exploration. Space exploration tasks are multifaceted and require a range of skills that no single robot may possess.
Building and maintaining a colony on Mars, for example, would require collaborative efforts from multiple robots with diverse capabilities, such as excavation, construction, maintenance, and repair. In such complex missions, traditional approaches may not be sufficient, while multi-robot collaboration through RL can play a vital role in achieving objectives. By enabling robots to work together in a way that mimics human teamwork, we can create systems that are capable of performing complex tasks in a more efficient and effective manner.

\noindent\textbf{Overview of related work.} Reinforcement learning (RL) has received much attention in many fields~\cite{guo2023backdoor,BaiZLZ21iros, siedler2022dynamic, zhang2020decentralized}. Recent research has demonstrated the potential benefits of multi-robot collaboration through RL in various real-world applications~\cite{BaiZLZ21iros, siedler2022dynamic, zhang2020decentralized}. For example, RL-based collaboration among drones in a reforestation task led to faster completion times and better overall performance~\cite{siedler2022dynamic}. Similarly, RL-based collaboration has been effective in object transportation scenarios~\cite{zhang2020decentralized}. These examples highlight the potential of multi-robot collaboration through RL to improve performance in real-world applications.

However, the emulation of human teamwork in multi-robot systems is not without its challenges. One such challenge lies in the occurrence of dilemmas often faced in human collaboration. As with human teams, robots may encounter situations where individual goals conflict with collective ones, or where coordination and cooperation come at a cost.
A particularly interesting perspective of multi-robot collaboration is social dilemmas, where individual robots face a trade-off between self-interests and collective benefits. These dilemmas can arise in various scenarios, such as task allocation, resource sharing, and navigation, and can lead to suboptimal outcomes, reduced efficiency, and conflicts~\cite{stimpson2003learning,yu2020distributed}. To address these challenges, several approaches, including game theory, machine learning, and communication strategies, have been proposed~\cite{10.5555/3237383.3237408, 10.5555/3495724.3496999, han2022solution, he2022robust}. For instance, an Admission-based hierarchical multi-agent RL strategy has been used to address cooperation and equality~\cite{app13031807}, while an adherence-based multi-agent reinforcement learning algorithm enhances the performance of RL agents by rewarding adherence to increase coordination and collaboration~\cite{yuan2022adherence}. These studies emphasize the importance of effective communication and coordination strategies to mitigate the effects of social dilemmas and improve overall performance in various applications, such as robotics, transportation, manufacturing, and environmental monitoring.

\noindent\textbf{This work.} Although recent research has investigated how communication can mitigate social dilemmas and improve performance in multi-robot systems, there is still much to learn about how environmental factors such as miscommunication and conflicting intentions can affect cooperation. Self-interested or even adversarial robots can further complicate collaboration, making it crucial to explore possible ways adversaries can manipulate the situation to achieve their intentions.

This paper examines how manipulating multi-robot communication can (either positively or negatively) impact the performance of multi-robot systems within social dilemmas. Specifically, we focus on two distinct forms of manipulation: policy and incentive  manipulation of the reward function. 
Through incentive reward manipulation, we control the flow of reward communication, in terms of both sending/receiving rewards to/from other robots; while through policy manipulation, we can interact with the environment with malicious policies to manipulate the behaviors.
Our analysis demonstrates the possibility of both positive and negative impacts on the task outcome. Positive impacts include faster convergence to the global optimum reward, which can guarantee to maximization of collective rewards for chosen robots. Conversely, negative impacts can reduce the success rate of the overall task performance. Our work thus introduces a new angle for manipulation in recent multi-agent reinforcement learning social dilemmas that utilize a unique reward function for incentivization.

To evaluate our proposed approach, we employ two social dilemma environments, the Escape room (ER)~\cite{10.5555/3495724.3496999} and Iterative Prisoners Dilemma (IPD)~\cite{10.5555/3237383.3237408}.
We use the Gazebo simulator to conduct experiments and demonstrate the effectiveness of our approach in these environments. 

Our contributions can be summarized as follows:

\begin{itemize}
\item This paper presents a novel approach to manipulating the collaborative behavior of multi-robot reinforcement learning (RL) systems in a social dilemma configuration by manipulating inter-robot communication. Two manipulation methods, referred to as policy and incentive reward manipulation, have been designed and implemented to effectively yield both negative and positive impacts on other agents and the final task results. 
\item The efficacy of our approach has been comprehensively evaluated in two different environments in Gazebo, and experimental results have been presented to demonstrate the impact of these manipulations on the performance of multi-robot systems, including 1) faster convergence to the global optimum, 2) maximizing collective reward for a chosen robot, and 3) reducing the success rate of task completion due to certain adversarial robots.
\item This work presents one of the first attempts to investigate the manipulation of multi-robot collaboration through RL and provides insights into the impact of manipulation on task performance in social dilemmas. The proposed approach has implications for various real-world applications, such as robotics, transportation, and manufacturing, where multi-robot collaboration is becoming increasingly important.
\end{itemize}

%% file: body/background.tex
To address the problem discussed in this paper, a clear understanding of how reinforcement learning can facilitate effective collaboration among multiple robots is crucial.

\subsection{Multi-Robot RL based on Policy Gradient}

To explain how policy gradient-based reinforcement learning (RL) works, we provide a concise description in an episodic, discrete-action setting. At each time step $t$, the agent observes the environment state $s_t \in S$, selects an action $a_t$ based on a policy $\pi$ from the action space $A$, and receives a reward $r_t$ from the environment. The agent's policy, represented by $\theta$, maps the state representation to a probability distribution over actions. The policy's value, denoted by $J(\pi_\theta)$ or $J(\theta)$, is the expected discounted sum of rewards obtained by the agent when following the policy $\pi_\theta$, and can be expressed as:

\begin{equation}
J(\theta)=E_\theta\left[\sum_{t=0}^{\infty} \gamma^t r_t\right]
\label{eq1}
\end{equation}

The gradient of the value $J$ with respect to the policy parameters $\theta$ can be calculated as follows \cite{NIPS1999_464d828b}: for all time steps $t$ within an episode,

\begin{equation}
\nabla_\theta J(\theta)=E_\theta\left[G\left(s_t, a_t\right) \nabla_\theta \log \pi_\theta\left(a_t \mid s_t\right)\right]
\label{policyGradient}
\end{equation}

Here, $G\left(s_t, a_t\right)=\sum_{i=t}^{\infty} \gamma^{i-t} r_i$ represents the return until termination.

\subsection{Incentivization in Policy Gradient}
Cooperative robotics requires effective incentivization methods to encourage agents to reach the optimal total reward in the environment. In multi-agent reinforcement learning (RL), each agent can have intrinsic and extrinsic sources of reward. While extrinsic rewards come from the environment, intrinsic rewards can be generated based on different policies~\cite{Zheng2018OnLI}. However, in cooperative tasks where the reward distribution varies per task, a self-interested approach may not be effective. In such cases, using an incentive function between robots can help them achieve the optimal total reward.
LIO~\cite{10.5555/3495724.3496999} proposes an approach where each agent learns an incentive function by considering its impact on the recipients' behavior and its own objective. LIO describes the general case of $N$ agents, where each agent has its own observation $o^i$, action $a^i$, and incentive function $r_{\eta^i}$ that maps its observation and all other agents' actions to rewards. The notation $i$ is used to refer to the reward-giving part of an agent and $j$ for the part that learns from received rewards. The notations $-i$ and $-j$ refer to all reward-givers/receivers except themselves. At each time step $t$, each recipient $j$ receives a total reward defined as:

\begin{equation} \label{lio_reward_func}
r^j\left(s_t, \mathbf{a}_t, \eta^{-j}\right):= \eqnmarkbox[WildStrawberry]{env}{r^{j, \mathrm{env}}\left(s_t, \mathbf{a}_t\right)}+\eqnmarkbox[Plum]{sigma}{\sum_{i \neq j} r_{\eta^i}^j\left(o_t^i, a_t^{-i}\right)}
\end{equation}
\annotate[yshift=-1em]{below,left}{env}{environment reward}
\annotate[yshift=0.5em]{above,left}{sigma}{Incentive reward}


%% file: body/design.tex
\begin{figure}
\centering
\includegraphics[width=2.6in]{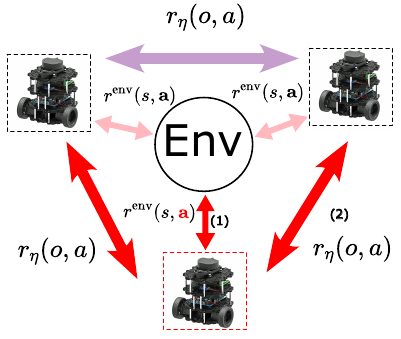}
\caption{ The red agent can manipulate through incentive function $(2)$ and policy manipulation $(1)$, as described in Eq.\ref{lio_reward_func}. By sending/receiving incentive rewards and taking malicious action based on certain policies, the red agent can manipulate other agents and affect game dynamics.}
\label{bigpic_fig}
\end{figure}

In the context of incentivized RL for social dilemmas, we are introducing two main forms of manipulation: incentive manipulation and policy manipulation, as shown in Figure~\ref{bigpic_fig}. A benign agent employs an iterative approach for bilevel optimization, in which the upper level optimizes the incentive function by considering the lower level's receivers' policy optimization. Using Eq.\ref{lio_reward_func} and the policy gradient of Eq.\ref{policyGradient}, we update the policy parameters $\theta^j$ with learning rate $\beta$ as follows:

\begin{equation}\label{newtheta}
\hat{\theta}^j \leftarrow \theta^j+\beta \left(  \nabla _{\theta^j} J^{ \pi }\left(\theta^j, \eta^{-j}\right)   \right)
\end{equation}

We then use the new parameter $\hat{\theta}^j$ to generate a new trajectory $\hat{\tau}^j$. Each reward-giver $i$ updates its individual incentive function parameters $\eta^i$ to maximize the following individual objective:

\begin{equation}\label{updateetha}
\nabla _{\eta^i} J^i\left(\hat{\tau}^i, \tau^i, \hat{\boldsymbol{\theta}}, \eta^i\right):=\mathbb{E}_{\hat{\boldsymbol{\pi}}}\left[\sum_{t=0}^T \gamma^t \hat{r}_t^{i, \text { env }}\right]-\alpha L\left(\eta^i, \tau^i\right)
\end{equation}
The first part of Eq.\ref{updateetha} is the expected extrinsic return of the reward-giver in the new trajectory $\hat{\tau}^i$, and the second term is a cost for giving rewards in the old trajectory $\tau^i$.

Next, we will provide a detailed explanation of the two manipulation methods of PIMbot.

\subsection{Incentive Reward Manipulation}

In multi-agent RL with incentivization setup, agents communicate by sending and receiving incentive rewards, which affects their actions and can lead to efficient cooperation. However, this communication channel is also a weak point that can be manipulated. In this section, we introduce two methods for conducting incentive reward manipulation.

\subsubsection{Partial Communication}

In a multi-agent RL setting, an agent acting in self-interest may choose to ignore incentive rewards and prioritize maximizing its own extrinsic reward. This may lead to the agent behaving differently from other agents, hindering collaboration and potentially resulting in suboptimal outcomes for the group as a whole. However, such behavior could still achieve the best possible outcome for the agent's self-interest.

To influence the behavior of other agents, an adversarial agent can manipulate the incentive communication by altering the second part of Eq.\ref{lio_reward_func}. Nevertheless, it still needs to provide incentive rewards to other agents using the same principle as Eq.\ref{updateetha}. We define self-centered reward $r^{SC}$ as:

\begin{equation} \label{adversaryreward}
r^{SC}\left(s_t, \mathbf{a}_t\right) := \eqnmarkbox[WildStrawberry]{env}{r^{\mathrm{env}}\left(s_t, \mathbf{a}_t\right)}
\end{equation}

This manipulation in the reward function will affect the update of $\theta^{Adv}$ to maximize reward in the environment, as shown in Eq.\ref{newtheta}.

\subsubsection{Fake Incentive Reward}
The incentive reward communication can be exploited by the adversarial agent to its advantage. By manipulating the incentive rewards it sends to other agents, the adversarial agent can induce them to behave in ways that benefit the adversarial agent's objectives. This can happen because there is no sanity check for receiving incentive rewards from other agents.

Specifically, the adversarial agent sends a constant positive number $C^{Adv} \in |\mathbb{N}|$ as the incentive reward. as mentioned in Eq.~\ref{Cinq}, $C^{Adv}$ should be much greater than what is available in the environment and provided by other agents to discourage other agents from taking any action.

\begin{equation} \label{Cinq}
C^{Adv} \gg r^{\mathrm{env}}\left(s_t, \mathbf{a}t\right),~ r_{\eta^i}^j\left(o_t^i, a_t^{-i}\right)
\end{equation}

For the reward function of other agents, we introduce this variation of Eq.\ref{lio_reward_func}:

\begin{equation} \label{fake_reward}
r^j\left(s_t, \mathbf{a}_t, \eta^{-j}\right):= \eqnmarkbox[WildStrawberry]{env}{r^{j, \mathrm{env}}\left(s_t, \mathbf{a}_t\right)}+\eqnmarkbox[Plum]{sigma}{\sum_{i \neq j} r_{\eta^i}^j\left(o_t^i, a_t^{-i}\right)} + \eqnmarkbox[RoyalPurple]{fake}{C^{Adv}} 
\end{equation}
\annotate[yshift=1.7em]{above,left}{fake}{Adv incentive reward}
\annotate[yshift=0.5em]{above,left}{sigma}{from all agents except Adv}

Since $C^{Adv}$ is dominating the other two parts of Eq.\ref{fake_reward}, the total reward received by agent $j$ is approximately equal to $C^{Adv}$. In other words, using the reward function in Eq.\ref{fake_reward} will result in no changes to the $J^{ \pi }\left(\theta^j, \eta^{-j}\right)$ of Eq.\ref{newtheta}, which leads to $\hat{\theta}^j = \theta^j$. This means that there will be no policy change and no exploration ($\nabla J^{ \pi } = 0$).

\subsection{Policy Manipulation}
In cooperative multi-agent tasks, such as social dilemma games,  all agents must work together to complete the task. However, an adversarial agent can play a significant role in the overall task by participating and taking appropriate actions based on its policy that can influence the behavior of other agents. Specifically, the adversarial agent may use policy manipulation to induce other agents to perform actions that benefit the adversarial agent's objectives. 

\subsubsection{Bypass Policy}

Social dilemma games are characterized by a group of agents working towards a shared goal, requiring each agent to take actions that benefit the group. However, conflicts may arise between an agent's self-interest and the need to cooperate for the good of the group.
This conflict can be exploited by an adversarial agent who manipulates the environmental dynamics. One way an adversarial agent can achieve this is by circumventing policies that prohibit an agent from taking actions or contributing to the group's effort. Such actions can have a significant impact on the overall outcome, as the final task cannot be completed without the adversarial agent's participation.
The adversarial agent can force other agents to act in ways that serve the adversarial agent's objectives by failing to take any action, thereby bypassing policies and exploiting the tension between individual self-interest and the collective good.

\begin{figure}[!t]
\centering
\includegraphics[width=3.5in]{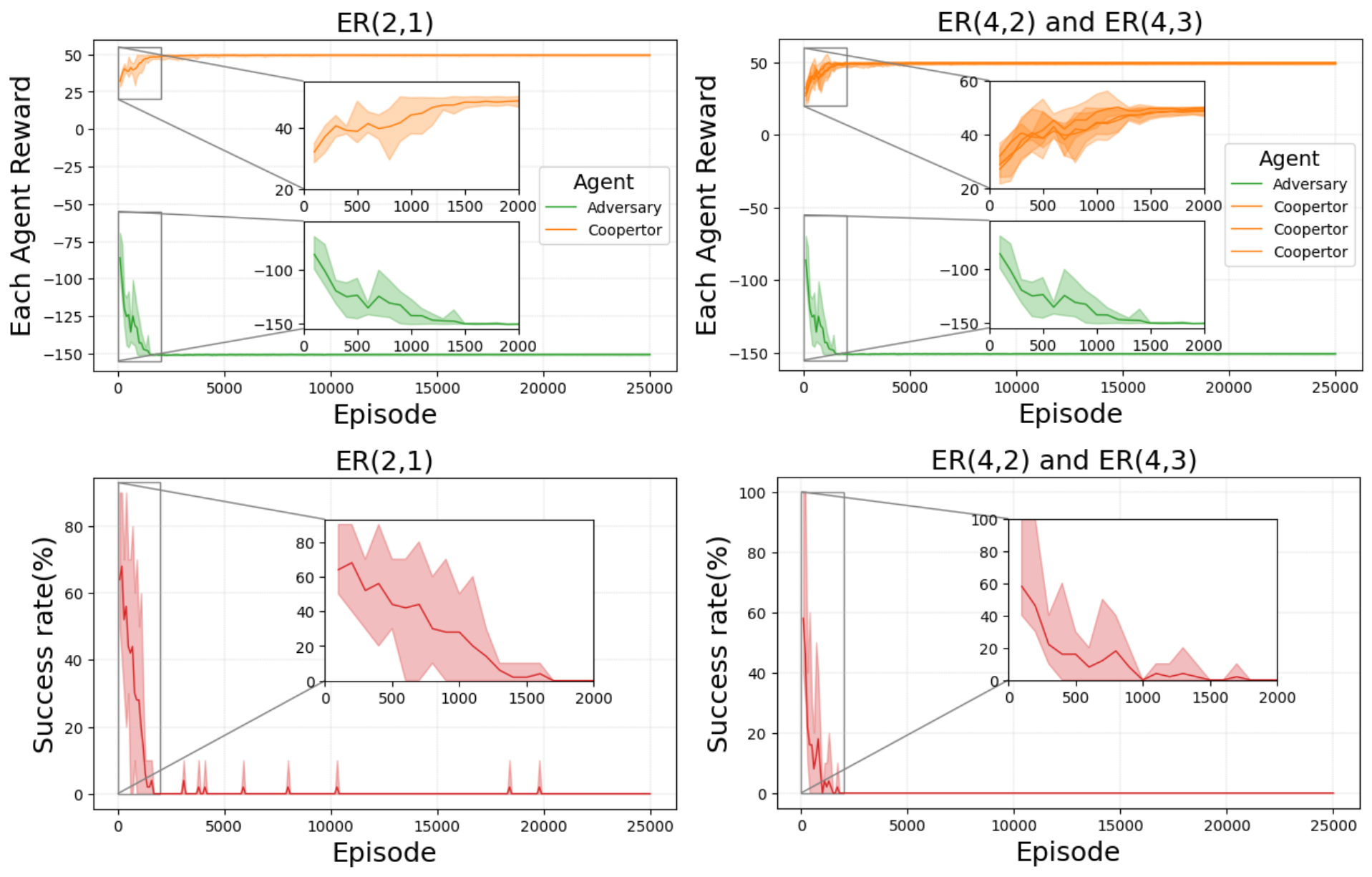}
\caption{Impact of fake incentive rewards as a manipulation method for incentive rewards. The use of fake incentive rewards leads to a lack of cooperation, as collaborators receive undeserved rewards, resulting in a significant drop in the success rate of task completion. The zoomed-in section shows the first 2000 episodes in which the agents are exploring the environment.}
\label{bignum}
\end{figure}

\subsubsection{Reverse Policy} 
In social dilemma games, all agents cooperate to achieve a shared goal using the same policies. However, an adversarial agent can leverage this commonality to manipulate the game by using a different policy to hinder the performance of benign agents. To achieve this, we introduce an agent that aims to minimize its own rewards in the environment. By doing so, the adversarial agent tries to deviate from solutions that benefit all agents and pursues its own objectives. Meanwhile, all benign agents aim to maximize their rewards using Eq. \ref{policyGradient}. They calculate their policy by using Eq. \ref{PGforgeniun}, which maximizes the expected sum of discounted rewards over time.

\begin{equation} \label{PGforgeniun}
\max _{\theta^j} J^{\pi}\left(\theta^j, \eta^{-j}\right):=\mathbb{E}_\theta\left[\sum_{t=0}^T \gamma^t r^j\left(s_t, \mathbf{a}_t, \eta^{-j}\right)\right]
\end{equation}

The adversarial agent, on the other hand, flips the rewards sign in Eq. \ref{PGforgeniun} to calculate $\min_{\theta^j} J^{\pi}\left(\theta^j, \eta^{-j}\right)$ and takes actions that minimize its own rewards. This has two sides of effect based on Eq. \ref{lio_reward_func}. First, the adversarial agent tries to minimize the extrinsic rewards provided in the game. Second, it takes actions that force other agents to send lower and lower incentive rewards. This manipulation by the adversarial agent can cause benign agents to act in ways that benefit the adversarial agent's objectives.

\begin{figure}[!t]
\centering
\includegraphics[width=3.5in]{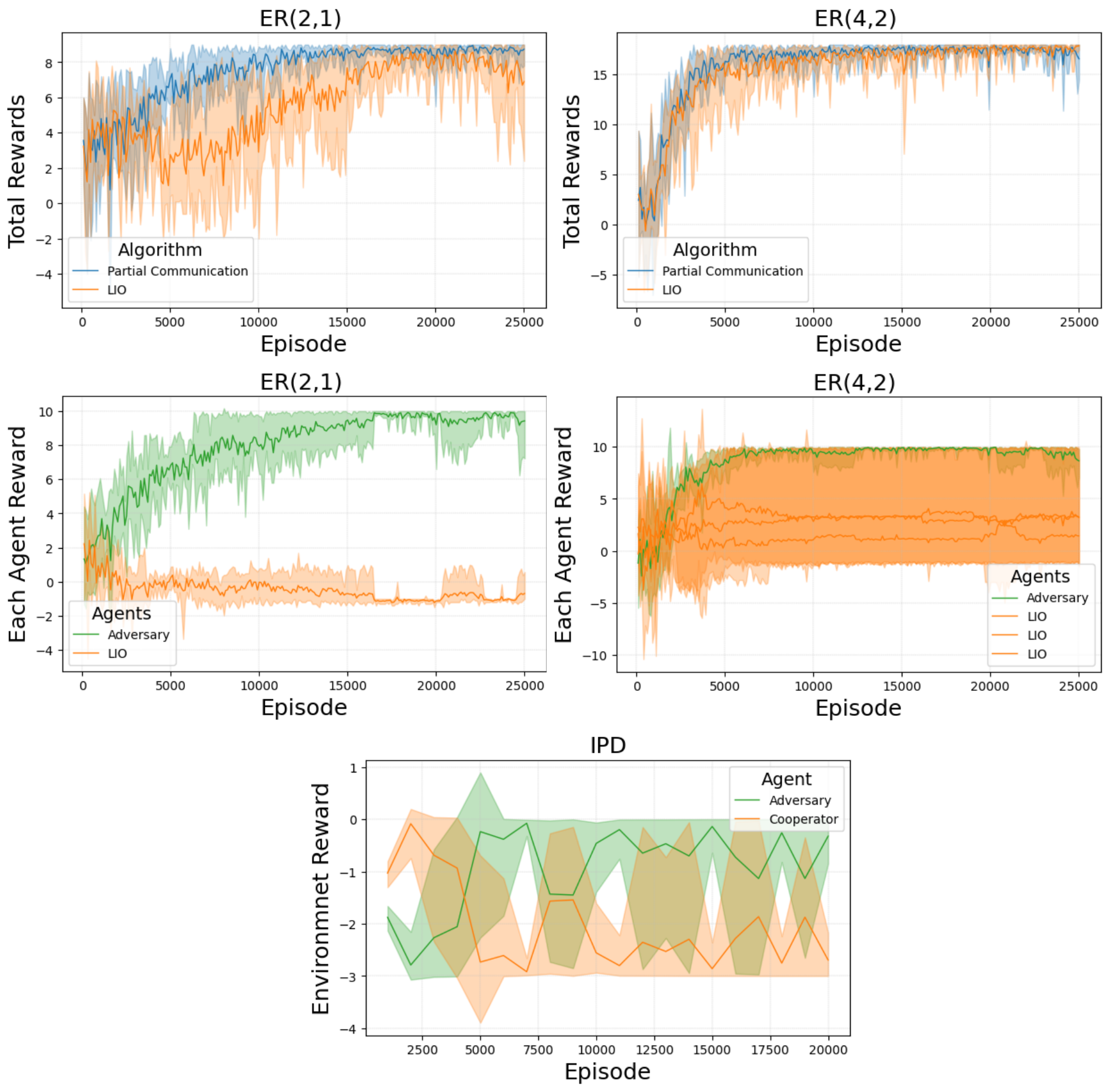}
\caption{Impact of partial communication on task completion time. The bottom plots depict the adversarial agent's ability to obtain the maximum reward in each run using partial communication in ER and IPD environments.}
\label{convergance}
\end{figure}

\begin{table}[]
\centering
\caption{Comparison of Convergence Time for Various Methods}
\label{table:conv}
\begin{tabular}{|c|c|c|}
\hline
\multirow{2}{*}{ Method } & \multicolumn{2}{c|}{ Convergence Time (episode$*10^3$ )} \\
\cline { 2 - 3 } & ER(4,2) & ER(2,1) \\
\hline LIO~\cite{10.5555/3495724.3496999} & $14$ & $20$ \\
Partial Communication & $10.5$ & $12.5$ \\
\hline
\end{tabular}
\end{table}

%% file: body/eval.tex
\begin{figure*}[!t]
\centering
\includegraphics[width=7in]{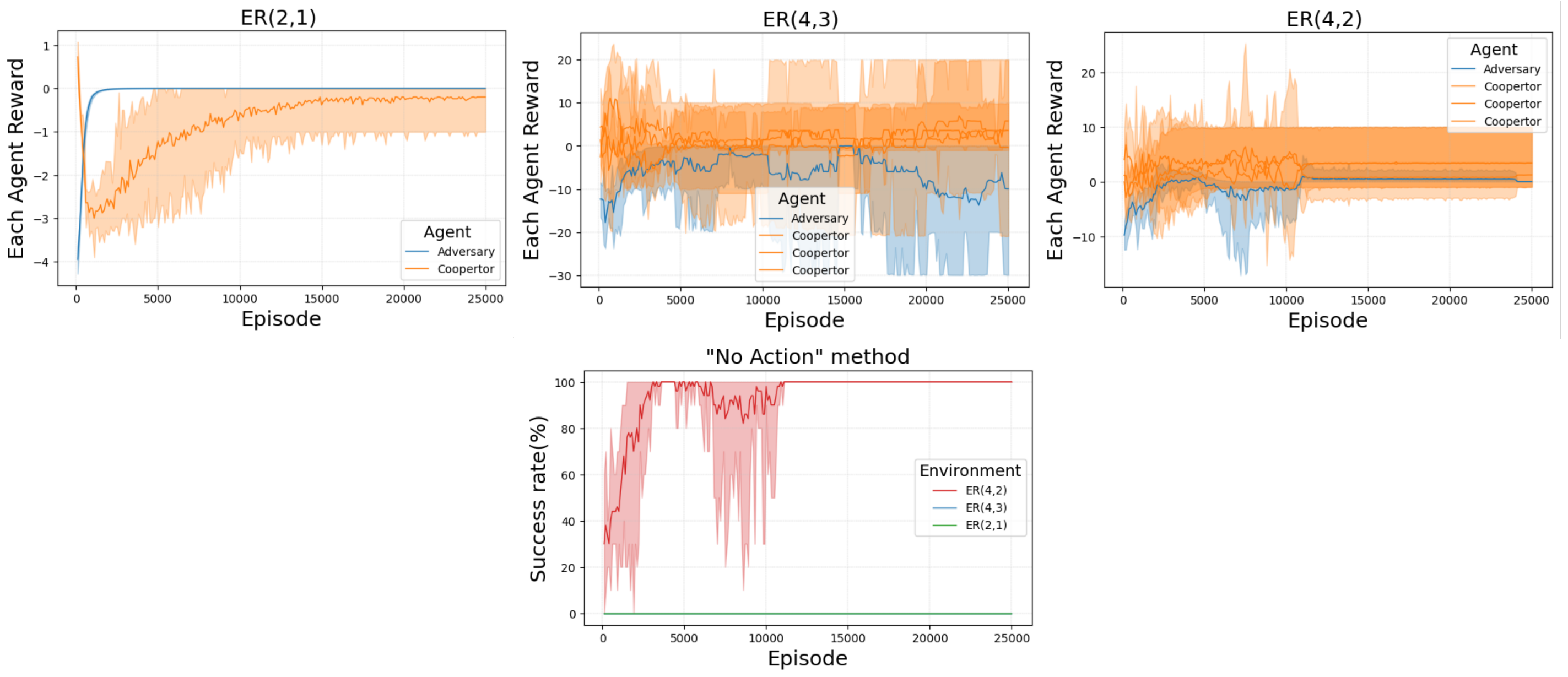}
\caption{Impact of the ''Bypass Policy'' method of policy manipulation in ER environments. Cooperators who send negative incentive rewards are attempting to incentivize the adversarial agent to take certain actions that align with the cooperative goal. The figure also demonstrates the concept of redundancy in the results of ER(4,2) and ER(4,3), where ER(4,2) achieves a high success rate by bypassing the adversarial agent with other cooperators, while the adversarial agent uses the method of policy manipulation.}
\label{noaction}
\end{figure*}

\subsection{Experimental Setup}
In this section, we will present the evaluation environment used to test our methods for incentive manipulation in multi-robot reinforcement learning within social dilemmas by manipulating the policy and incentive rewards. We present the results of 10 separate runs for each environment to demonstrate the effectiveness of PIMbot's method.

\subsubsection{Escape Room (ER)}
In our evaluation, we utilized Gazebo to implement the Escape Room environment \cite{10.5555/3495724.3496999}. The Escape Room environment ER($N, M$) is a Markov game for N players with individual extrinsic rewards. To successfully exit the environment, an agent must receive a +10 extrinsic reward by exiting a door. However, the door can only be opened when $M$ other agents cooperate to pull the lever ($M<N$), which incurs an extrinsic penalty of -1 for any movement, thus discouraging all agents from taking cooperative action. We conducted experiments with the cases of ($N=2$, $M=1$), ($N=4$, $M=2$), and ($N=4$, $M=3$) to evaluate the effectiveness of our method. 

\subsubsection{Iterated Prisoner's Dilemma (IPD)}
In addition to our evaluation of the Escape Room environment, we conducted tests on the memory-1 Iterated Prisoner's Dilemma environment using our manipulation method \cite{10.5555/3237383.3237408}. In this environment, each agent observes the joint action taken by themselves and the other agent in the previous round. We used the extrinsic reward table for the environment, to evaluate the performance of our method. Table~\ref{table:IPD_reward} specifies the extrinsic rewards for each possible outcome of the environment, including mutual cooperation, mutual defection, and unilateral defection.

\begin{table}[H] \label{IPD_reward}
\centering
\caption{IPD reward table}
\label{table:IPD_reward}
\begin{tabular}{c|cc}
$\mathrm{Agent} 1 / \mathrm{Agent} 2$ & $\mathrm{Cooperate (C)}$ & $\mathrm{Defect (D)}$ \\
\hline $\mathrm{Cooperate (C)}$ & $(-1,-1)$ & $(-3,0)$ \\
$\mathrm{Defect (D)}$ & $(0,-3)$ & $(-2,-2)$
\end{tabular}
\end{table}

\subsection{Implementation Details}
To enable a fair comparison between our approach and LIO~\cite{10.5555/3495724.3496999}, we used the same hyperparameters $\theta$ and $\eta$ for Eq.\ref{newtheta} and Eq.\ref{updateetha}. For our experiments, we deployed the Escape Room (ER) environment in both ROS and Gazebo for two and four-player versions of the environment. We used the TurtleBot3 simulation platform to demonstrate the behavior of agents in the ER environment. Both the simulation and manipulation were run on an Intel Core i7-10700K CPU.


\subsection{Experimental Results}
To demonstrate the effectiveness of the manipulation methods discussed in the previous section for multi-robot reinforcement learning in social dilemmas, we present the results of implementing these methods in both environments in this section. Specifically, we will discuss the results of each manipulation method used in the experiments. By evaluating the performance of our approach in comparison to the baseline LIO~\cite{10.5555/3495724.3496999}, we can show the benefits of our proposed manipulation techniques.

\begin{figure*}[!t]
\centering
\includegraphics[width=7in]{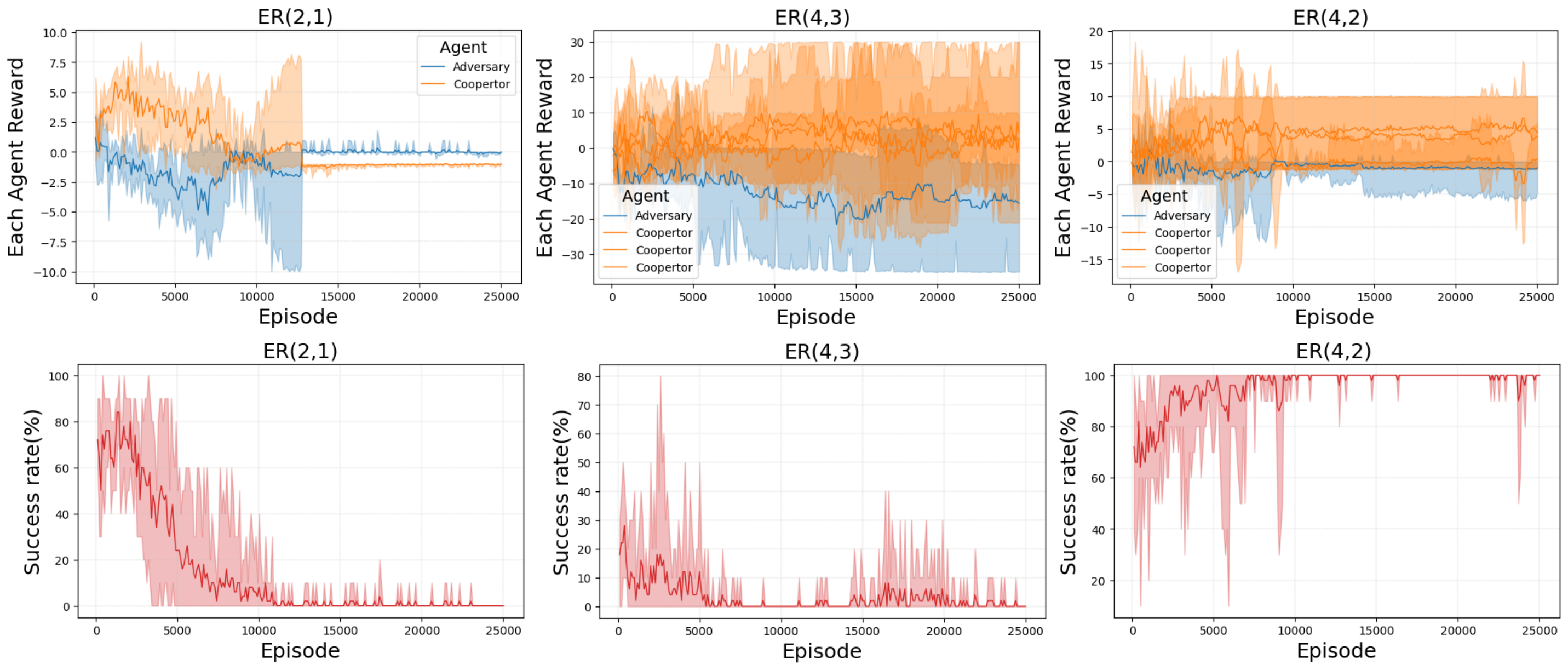}
\caption{Impact of an adversarial agent's use of the "Reverse Policy" method to manipulate policy. It demonstrates that the adversarial agent's attempts to minimize the collective reward leads to a decrease in the success rate. The redundancy between ER(4,2) and ER(4,3) is visible in the figure, as shown in the bottom middle and bottom right. Despite manipulating both environments, the success rate remains low in ER(4,3) (bottom middle).}
\label{a2d}
\end{figure*}

\subsubsection{Incentive Reward Manipulation}
In this section, we discuss the methods used for incentive manipulation, namely partial communication and fake incentive rewards, which exploit the incentive channel between agents to achieve their objectives. 

The partial communication method controls the direction of the incentive function and thus maximizes the adversarial agent's extrinsic reward. As demonstrated in Figure~\ref{convergance}, the adversarial agent consistently achieves greater reward values than other benign agents in ER(2,1), ER(4,2), and IPD.

Although the adversarial agent's goal is to maximize its own reward function, our approach enables the entire team of agents to solve the task in a shorter period of time. As shown in the results of ER(2,1) and ER(4,2) in Figure~\ref{convergance}, our approach reaches the global optimum reward faster than LIO in 37.5\% for ER(2,1) and 25\% in ER(4,2), as mentioned in Table~\ref{table:conv}.

The objective of reaching the optimal point in the IPD environment differs from that in the ER environment. In IPD, the global optimal reward is achieved when all agents reach their individual maximum reward function, while in ER, some agents must gain less reward to assist others in reaching their maximum reward and completing the task. Our proposed method enabled an agent to reach the maximum reward in the IPD environment. However, an adversarial agent may only maximize its own extrinsic reward and cannot assist in achieving the total optimal reward for both agents. Therefore, utilizing our method with an adversarial agent may not result in the same outcomes.

In the Fake Incentive Reward manipulation method, the adversarial agent sends out malicious rewards to other agents using the incentive channel. In the Escape Room environment, we set the value of $C^{Adv}$ to 50 in Eq.\ref{fake_reward}. The reason for choosing 50 as the fake reward is that it is greater than the maximum reward that agents can receive for taking any action in the environment.
As shown in the results presented in Figure~\ref{bignum}, the agents explore the environment for 1800 episodes, which is why they achieve some success rate during this period.

\subsubsection{Policy Manipulation}
In the environments discussed previously, we applied the policy manipulation method, which enables the adversarial agent to manipulate the environment by altering the action caused by its policy. The following section presents the results of both the ''Bypass Policy'' and ''Reverse Policy'' methods. 

We evaluated the ''Bypass Policy'' method as the first approach to policy manipulation in the ER environment. In this method, the agent bypasses the policy and as a result, it does not take any action and remains in its initial state. However, since social dilemmas require the cooperation and actions of all agents to reach the optimal result, this method hinders the achievement of the optimal result. The results presented in Figure~\ref{noaction} show that for ER(4,3) and ER(2,1), the agents failed to reach the optimal reward, and the success rate for those environments is zero.

The second method of action manipulation is the ''Reverse Policy'' method, where the adversarial agent uses a different policy than other benign agents. The adversarial agent is minimizing the collective reward by changing its policy from gradient ascent to gradient descent. The adversarial agent takes actions with minimum reward and incentivizes other agents to give it the minimum incentive reward possible. The effectiveness of this method is shown in Figure~\ref{a2d}, where the success rate of task completion drops to zero.

One important observation is that environments can be classified based on whether all agents are required to collaborate to accomplish the task or whether some agents can independently contribute to achieving the goal. In this regard, we introduce the redundancy term in environments to analyze the effect of adversarial manipulation on the task completion process. When the adversarial agent abandons the main task, other agents still try to maximize their rewards. In redundant environments, other agents will ignore the adversarial agent's action and continue to move toward finishing the task. Yet, in non-redundant environments, all agents require the activity of the adversarial agent to finish the task, and they will eventually give up. In the case of our experiments, the environments of ER(4,3) and ER(2,1) are non-redundant, making the policy manipulation method successful in reducing the success rate and preventing the attainment of the total optimal reward. However, in the environment of ER(4,2), where two agents are required to pull the lever and only one agent is needed to reach the door, the adversarial agent's abandonment of other agents does not affect the final task completion process. As shown in Figure~\ref{a2d} and Figure~\ref{noaction}, agents still reach the total optimal global reward.

%% file: body/discussion.tex
We have explored potential methods to manipulate dilemmas in multi-robot collaboration. However, an essential aspect of unveiling problems is the development of strategies to prevent them.
In the context of this paper, one potential safeguard against the challenges posed by our approach could be for the agents to maintain a certain degree of exploration that is not dependent on collaboration protocols. However, this strategy may not be effective in situations with no optimal solution.
Alternatively, a central unit could manage policies and reward exchanges between robots. This would ensure task order and possibly avoid some dilemmas.
For problems that lack optimal solutions, it may be advantageous to reconfigure them, if possible, to rely on fewer robots. This could be a strategy to reduce complexity and potential sources of conflicts.
These potential solutions open up new research avenues that need further exploration.

%% file: body/conclusion.tex
In conclusion, this study has revealed that in multi-robot reinforcement learning collaborations, robots can have different intentions when faced with social dilemmas. By introducing an adversarial agent into two social dilemma environments and manipulating its policy and incentive reward channel, we were able to achieve three main goals. Firstly, the adversarial robot successfully obtained the maximum reward, highlighting the effectiveness of its malicious behavior. Secondly, the adversarial robot was able to converge faster to the optimum reward, demonstrating the potential impact of its behavior on the overall performance of the multi-robot system. Finally, the adversarial robot cooperated maliciously, causing issues for the other agents. These findings emphasize the need for further research into developing secure, robust, and equitable methods for multi-robot collaborations in the presence of adversarial agents. Ultimately, this work provides valuable insights into the challenges and opportunities involved in using multi-robot systems to address complex real-world problems.

%% file: body/ack.tex
We would like to express our gratitude to Jiachen Yang for the insightful discussions concerning this paper.

%% file: main.bbl
\begin{thebibliography}{10}
\providecommand{\url}[1]{#1}
\csname url@rmstyle\endcsname
\providecommand{\newblock}{\relax}
\providecommand{\bibinfo}[2]{#2}
\providecommand\BIBentrySTDinterwordspacing{\spaceskip=0pt\relax}
\providecommand\BIBentryALTinterwordstretchfactor{4}
\providecommand\BIBentryALTinterwordspacing{\spaceskip=\fontdimen2\font plus
\BIBentryALTinterwordstretchfactor\fontdimen3\font minus
  \fontdimen4\font\relax}
\providecommand\BIBforeignlanguage[2]{{%
\expandafter\ifx\csname l@#1\endcsname\relax
\typeout{** WARNING: IEEEtran.bst: No hyphenation pattern has been}%
\typeout{** loaded for the language `#1'. Using the pattern for}%
\typeout{** the default language instead.}%
\else
\language=\csname l@#1\endcsname
\fi
#2}}

\bibitem{ahangar2019design}
S.~Ahangar, M.~V. Mehrabani, A.~P. Shorijeh, and M.~T. Masouleh, ``Design a
  3-dof delta parallel robot by one degree redundancy along the conveyor axis,
  a novel automation approach,'' in \emph{2019 5th Conference on Knowledge
  Based Engineering and Innovation (KBEI)}.\hskip 1em plus 0.5em minus
  0.4em\relax IEEE, 2019, pp. 413--418.

\bibitem{su15107790}
\BIBentryALTinterwordspacing
B.~Mehralizadeh, B.~Baradaran, S.~Nikkhoo, P.~Soleiman, and H.~Moradi, ``A
  sensorized toy car for autism screening using multi-modal features,''
  \emph{Sustainability}, vol.~15, no.~10, 2023. [Online]. Available:
  \url{https://www.mdpi.com/2071-1050/15/10/7790}
\BIBentrySTDinterwordspacing

\bibitem{TereshchukSBPDB19}
\BIBentryALTinterwordspacing
V.~Tereshchuk, J.~Stewart, N.~Bykov, S.~Pedigo, S.~Devasia, and A.~G. Banerjee,
  ``An efficient scheduling algorithm for multi-robot task allocation in
  assembling aircraft structures,'' \emph{{IEEE} Robotics Autom. Lett.},
  vol.~4, no.~4, pp. 3844--3851, 2019. [Online]. Available:
  \url{https://doi.org/10.1109/LRA.2019.2929983}
\BIBentrySTDinterwordspacing

\bibitem{AgrawalABM22iros}
\BIBentryALTinterwordspacing
A.~Agrawal, S.~H. Arul, A.~S. Bedi, and D.~Manocha, ``{DC-MRTA:} decentralized
  multi-robot task allocation and navigation in complex environments,'' in
  \emph{{IEEE/RSJ} International Conference on Intelligent Robots and Systems,
  {IROS} 2022, Kyoto, Japan, October 23-27, 2022}.\hskip 1em plus 0.5em minus
  0.4em\relax {IEEE}, 2022, pp. 11\,711--11\,718. [Online]. Available:
  \url{https://doi.org/10.1109/IROS47612.2022.9981353}
\BIBentrySTDinterwordspacing

\bibitem{GaoWZYWXWLXG22iros}
\BIBentryALTinterwordspacing
Y.~Gao, Y.~Wang, X.~Zhong, T.~Yang, M.~Wang, Z.~Xu, Y.~Wang, Y.~Lin, C.~Xu, and
  F.~Gao, ``Meeting-merging-mission: {A} multi-robot coordinate framework for
  large-scale communication-limited exploration,'' in \emph{{IEEE/RSJ}
  International Conference on Intelligent Robots and Systems, {IROS} 2022,
  Kyoto, Japan, October 23-27, 2022}.\hskip 1em plus 0.5em minus 0.4em\relax
  {IEEE}, 2022, pp. 13\,700--13\,707. [Online]. Available:
  \url{https://doi.org/10.1109/IROS47612.2022.9981544}
\BIBentrySTDinterwordspacing

\bibitem{ZhangQQXWZWC0ZL22iros}
\BIBentryALTinterwordspacing
Q.~Zhang, R.~Quan, S.~Qimuge, P.~Xia, J.~Wang, X.~Zan, F.~Wang, C.~Chen,
  Q.~Wei, H.~Zhao, X.~Liu, and F.~Qiao, ``{OCTOANTS:} {A} heterogeneous
  lightweight intelligent multi-robot collaboration system with
  resource-constrained iot devices,'' in \emph{{IEEE/RSJ} International
  Conference on Intelligent Robots and Systems, {IROS} 2022, Kyoto, Japan,
  October 23-27, 2022}.\hskip 1em plus 0.5em minus 0.4em\relax {IEEE}, 2022,
  pp. 2556--2563. [Online]. Available:
  \url{https://doi.org/10.1109/IROS47612.2022.9982135}
\BIBentrySTDinterwordspacing

\bibitem{guo2023backdoor}
J.~Guo, A.~Li, and C.~Liu, ``Backdoor detection and mitigation in competitive
  reinforcement learning,'' 2023.

\bibitem{BaiZLZ21iros}
\BIBentryALTinterwordspacing
R.~Bai, R.~Zheng, M.~Liu, and S.~Zhang, ``Multi-robot task planning under
  individual and collaborative temporal logic specifications,'' in
  \emph{{IEEE/RSJ} International Conference on Intelligent Robots and Systems,
  {IROS} 2021, Prague, Czech Republic, September 27 - Oct. 1, 2021}.\hskip 1em
  plus 0.5em minus 0.4em\relax {IEEE}, 2021, pp. 6382--6389. [Online].
  Available: \url{https://doi.org/10.1109/IROS51168.2021.9636287}
\BIBentrySTDinterwordspacing

\bibitem{siedler2022dynamic}
P.~D. Siedler, ``Dynamic collaborative multi-agent reinforcement learning
  communication for autonomous drone reforestation,'' \emph{arXiv preprint
  arXiv:2211.15414}, 2022.

\bibitem{zhang2020decentralized}
L.~Zhang, Y.~Sun, A.~Barth, and O.~Ma, ``Decentralized control of multi-robot
  system in cooperative object transportation using deep reinforcement
  learning,'' \emph{IEEE Access}, vol.~8, pp. 184\,109--184\,119, 2020.

\bibitem{stimpson2003learning}
J.~L. Stimpson and M.~A. Goodrich, ``Learning to cooperate in a social dilemma:
  A satisficing approach to bargaining,'' in \emph{ICML}.\hskip 1em plus 0.5em
  minus 0.4em\relax Citeseer, 2003, pp. 728--735.

\bibitem{yu2020distributed}
C.~Yu, Y.~Dong, Y.~Li, and Y.~Chen, ``Distributed multi-agent deep
  reinforcement learning for cooperative multi-robot pursuit,'' \emph{The
  Journal of Engineering}, vol. 2020, no.~13, pp. 499--504, 2020.

\bibitem{10.5555/3237383.3237408}
J.~Foerster, R.~Y. Chen, M.~Al-Shedivat, S.~Whiteson, P.~Abbeel, and
  I.~Mordatch, ``Learning with opponent-learning awareness,'' in
  \emph{Proceedings of the 17th International Conference on Autonomous Agents
  and MultiAgent Systems}, ser. AAMAS '18.\hskip 1em plus 0.5em minus
  0.4em\relax Richland, SC: International Foundation for Autonomous Agents and
  Multiagent Systems, 2018, p. 122–130.

\bibitem{10.5555/3495724.3496999}
J.~Yang, A.~Li, M.~Farajtabar, P.~Sunehag, E.~Hughes, and H.~Zha, ``Learning to
  incentivize other learning agents,'' in \emph{Proceedings of the 34th
  International Conference on Neural Information Processing Systems}, ser.
  NIPS'20.\hskip 1em plus 0.5em minus 0.4em\relax Red Hook, NY, USA: Curran
  Associates Inc., 2020.

\bibitem{han2022solution}
S.~Han, S.~Su, S.~He, S.~Han, H.~Yang, and F.~Miao, ``What is the solution for
  state adversarial multi-agent reinforcement learning?'' \emph{arXiv preprint
  arXiv:2212.02705}, 2022.

\bibitem{he2022robust}
S.~He, Y.~Wang, S.~Han, S.~Zou, and F.~Miao, ``A robust and constrained
  multi-agent reinforcement learning framework for electric vehicle amod
  systems,'' \emph{arXiv preprint arXiv:2209.08230}, 2022.

\bibitem{app13031807}
\BIBentryALTinterwordspacing
T.~Guo, Y.~Yuan, and P.~Zhao, ``Admission-based reinforcement-learning
  algorithm in sequential social dilemmas,'' \emph{Applied Sciences}, vol.~13,
  no.~3, 2023. [Online]. Available:
  \url{https://www.mdpi.com/2076-3417/13/3/1807}
\BIBentrySTDinterwordspacing

\bibitem{yuan2022adherence}
Y.~Yuan, T.~Guo, P.~Zhao, and H.~Jiang, ``Adherence improves cooperation in
  sequential social dilemmas,'' \emph{Applied Sciences}, vol.~12, no.~16, p.
  8004, 2022.

\bibitem{NIPS1999_464d828b}
\BIBentryALTinterwordspacing
R.~S. Sutton, D.~McAllester, S.~Singh, and Y.~Mansour, ``Policy gradient
  methods for reinforcement learning with function approximation,'' in
  \emph{Advances in Neural Information Processing Systems}, S.~Solla, T.~Leen,
  and K.~M\"{u}ller, Eds., vol.~12.\hskip 1em plus 0.5em minus 0.4em\relax MIT
  Press, 1999. [Online]. Available:
  \url{https://proceedings.neurips.cc/paper/1999/file/464d828b85b0bed98e80ade0a5c43b0f-Paper.pdf}
\BIBentrySTDinterwordspacing

\bibitem{Zheng2018OnLI}
Z.~Zheng, J.~Oh, and S.~Singh, ``On learning intrinsic rewards for policy
  gradient methods,'' in \emph{Neural Information Processing Systems}, 2018.

\end{thebibliography}
